\documentclass[11pt,a4paper]{article}
\pdfoutput=1
\usepackage[hyperref]{acl2021}
\usepackage{times}
\usepackage{latexsym}

\usepackage{microtype}

\aclfinalcopy % Uncomment this line for the final submission
\def\aclpaperid{1998} %  Enter the acl Paper ID here

\usepackage{graphicx}
\usepackage{latexsym}
\usepackage{amsmath,amssymb} % define this before the line numbering.
\usepackage{xcolor}
\usepackage{booktabs}
\usepackage{subcaption}
\usepackage{multirow}

\title{{\color{blue}GEM}: A {\color{blue}G}eneral {\color{blue}E}valuation Benchmark for {\color{blue}M}ultimodal Tasks}

\author{Lin Su$^1$ \quad Nan Duan$^2$ \quad Edward Cui$^1$ \quad Lei Ji$^2$ \quad Chenfei Wu$^2$ \quad Huaishao Luo$^3$ \\ \textbf{Yongfei Liu$^4$ \quad Ming Zhong$^1$ \quad Taroon Bharti$^1$ \quad Arun Sacheti$^5$} \\
$^1$ Bing Multimedia Team, Microsoft, China\\
$^2$ Natural Language Computing, Microsoft Research Asia, China\\
$^3$ Southwest Jiaotong University, China $^4$ ShanghaiTech University, China\\
$^5$ Bing Multimedia Team, Microsoft, United States\\
{\tt\small \{lins, nanduan, edwac, leiji, chewu, minzhon, tbharti, aruns\}@microsoft.com}\\
{\tt\small huaishaoluo@gmail.com,} {\tt\small liuyf3@shanghaitech.edu.cn}
}

\date{}

\begin{document}
\maketitle
\begin{abstract}
In this paper, we present \textbf{GEM}\footnote{https://github.com/microsoft/GEM} as a \textbf{G}eneral \textbf{E}valuation benchmark for \textbf{M}ultimodal tasks. Different from existing datasets such as GLUE \cite{DBLP:journals/corr/abs-1804-07461}, SuperGLUE \cite{DBLP:journals/corr/abs-1905-00537}, XGLUE \cite{liang2020xglue} and XTREME \cite{hu2020xtreme} that mainly focus on natural language tasks, GEM is a large-scale vision-language benchmark, which consists of \textbf{GEM-I} for image-language tasks and \textbf{GEM-V} for video-language tasks. Comparing with existing multimodal datasets such as MSCOCO \cite{chen2015microsoft} and Flicker30K \cite{vinyals2015tell} for image-language tasks, YouCook2 \cite{ZhXuCoCVPR18} and MSR-VTT \cite{xu2016msr-vtt} for video-language tasks, GEM is not only the largest vision-language dataset covering image-language tasks and video-language tasks at the same time, but also labeled in multiple languages. We also provide two baseline models for this benchmark. We will release the dataset, code and baseline models, aiming to advance the development of multilingual multimodal research.
\end{abstract}

\section{Introduction}

In recent years, large-scale pre-training has become the new paradigm in the natural language processing (NLP) field. These models have demonstrated surprisingly good generalization abilities and can be applied to different downstream tasks by a simple fine-tuning. Several comprehensive benchmarks are constructed to evaluate such powerful models, including GLUE \cite{DBLP:journals/corr/abs-1804-07461} and SuperGLUE \cite{DBLP:journals/corr/abs-1905-00537} for evaluating monolingual natural language understanding systems, XGLUE \cite{liang2020xglue} and XTREME \cite{hu2020xtreme} for evaluating multilingual natural language understanding and generation systems. Such pre-trained models have also been extended to vision-language scenarios \citep{lu2019vilbert,chen2019uniter,li2019unicodervl,li2020oscar,ni2021m3p,sun2019videobert,sun2019cbt,luo2020univl} to handle multimodal tasks such as image(or video)-text retrieval and image (or video) captioning. However, there is still no comprehensive benchmark dataset for evaluating such multimodal pre-trained models. Besides, most existing vision-language datasets are labeled in English only, which cannot be used to evaluate the qualities of such models on other languages. 

Motivated by this, we present \textbf{GEM}, a \textbf{G}eneral \textbf{E}valuation benchmark for \textbf{M}ultimodal tasks. Comparing with GLUE, SuperGLUE, XGLUE and XTREME, GEM is designed for evaluating the generalization capabilities of vision-language models and consists of two subsets: GEM-I, which evaluates text-to-image retrieval and image captioning capabilities, and GEM-V, which evaluates text-to-video retrieval and video captioning capabilities. Besides, it is also a multilingual dataset, where the natural language contexts are collected from a commercial search engine. We describe two vision-language pre-trained models, M$^3$P \cite{ni2021m3p} and m-UniVL, as the baselines for GEM-I and GEM-V, respectively, where M$^3$P is an existing multilingual image-language pre-trained model, m-UniVL is a multilingual extension of UniVL \cite{luo2020univl} for multilingual video-language tasks. The evaluation results of these two models on GEM are reported in the experiment part.

The key contribution of this paper is twofold: (1) we build GEM as the first large-scale multilingual multimodal benchmark, which can be used to evaluate the generalization capabilities of vision-language pre-trained models on a set of diversified multimodal tasks. (2) we provide two multilingual multimodal pre-trained models, M$^3$P and m-UniVL, as the baselines of GEM for image-language and video-language tasks, respectively. We hope GEM can further advance the research in the multimodal community, just as its predecessors did in the NLP community. 

\section{Dataset Construction}
\label{sec:dataset-construction}

\begin{figure*} [tp]
    \centering
    \includegraphics[width=\textwidth,height=\textheight,keepaspectratio]{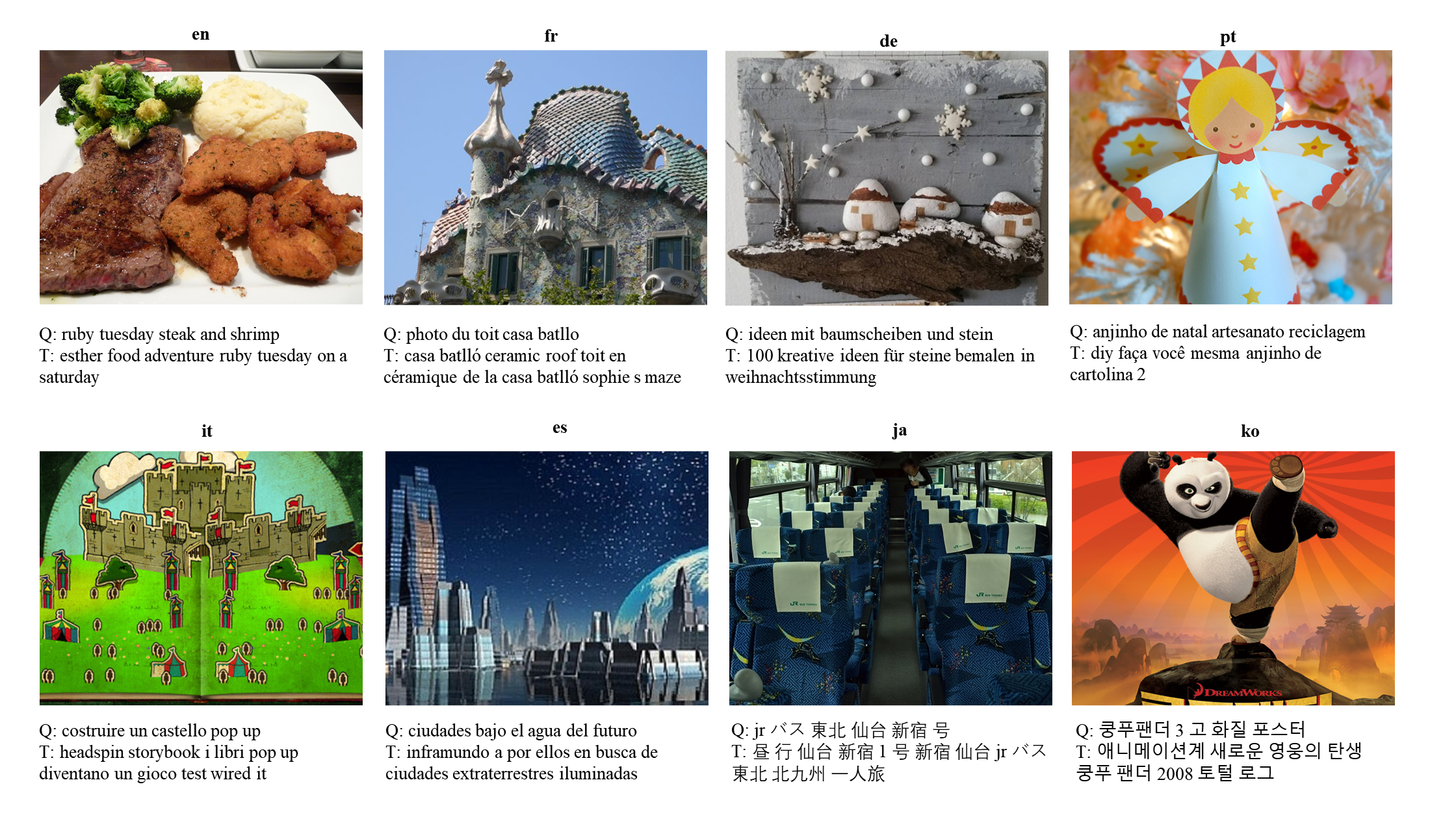}
    \caption{Examples in GEM-I dataset: Q: Query, T: Title.}
    \label{GEMI-case}
\end{figure*}

%Unlike existing multilingual text-image datasets, such as MSCOCO \cite{chen2015microsoft} or Multi30K \cite{elliott2016multi30k,elliott2017findings} that only contain three or four languages, the GEM-I dataset contains 1.2 million images with text descriptions in 20 languages. Moreover, the existing text-video datasets, such as Youcook2 \cite{ZhXuCoCVPR18} or MSR-VTT \cite{xu2016msr-vtt}, only contain text descriptions in English. However, our GEM-V dataset contains 99k videos (total video length xx hours) with text descriptions in as many as 30 languages. The text descriptions in GEM-I and GEM-V can be used for image retrieval or video retrieval, and image captioning or video captioning tasks. For each image or video, in addition to the text description, we also provide a short context, that can be used as additional information in both understanding and generation tasks. We will introduce more details about the usage of the context in our ablation study section.

\begin{table}
\scalebox{0.7}{
\centering
\begin{tabular}{lrrrr}
\hline \textbf{Language} & \textbf{Train} & \textbf{Dev} & \textbf{Test} & \textbf{Total} \\ \hline
English (en) & 998,000 & 1,000 & 1,000 & 1,000,000 \\
Spanish (es) & 18,000 & 1,000 & 1,000 & 20,000 \\
French (fr) & 18,000 & 1,000 & 1,000 & 20,000 \\
Italian (it) & 18,000 & 1,000 & 1,000 & 20,000 \\
Portuguese (pt) & 18,000 & 1,000 & 1,000 & 20,000 \\
German (de) & 18,000 & 1,000 & 1,000 & 20,000 \\
Korean (ko) & 8,000 & 1,000 & 1,000 & 10,000 \\
Polish (pl) & 8,000 & 1,000 & 1,000 & 10,000 \\
Catalan (ca) & 2,000 & 1,000 & 1,000 & 4,000 \\
Dutch (nl) & 2,000 & 1,000 & 1,000 & 4,000 \\
Japanese (ja) & 2,000 & 1,000 & 1,000 & 4,000 \\
Indonesian (id) & 2,000 & 1,000 & 1,000 & 4,000 \\
Vietnamese (vi) & 2,000 & 1,000 & 1,000 & 4,000 \\
Czech (cs) & 2,000 & 1,000 & 1,000 & 4,000 \\
Romanian (ro) & 2,000 & 1,000 & 1,000 & 4,000 \\
Turkish (tr) & 0 & 0 & 1,000 & 1,000 \\
Galician (gl) & 0 & 0 & 1,000 & 1,000 \\
Croatian (hr) & 0 & 0 & 1,000 & 1,000 \\
Hungarian (hu) & 0 & 0 & 1,000 & 1,000 \\
Malay (ms) & 0 & 0 & 1,000 & 1,000 \\
\hline
\textbf{Total} & \textbf{1,118,000} & \textbf{15,000} & \textbf{20,000} & \textbf{1,153,000} \\
\hline
\end{tabular}
}
\caption{\label{GEMI-table} Language distribution and data statistics of GEM-I for multilingual image-language tasks.}
\end{table}

\begin{table}
\scalebox{0.7}{
\centering
\begin{tabular}{lrrrr}
\hline \textbf{Language} & \textbf{Train} & \textbf{Dev} & \textbf{Test} & \textbf{Total} \\ \hline
German (de) & 3,316 & 1,000 & 1,000 & 5,316 \\
Portuguese (pt) & 3,258 & 1,000 & 1,000 & 5,258 \\
Dutch (nl) & 2,961 & 1,000 & 1,000 & 4,961 \\
Spanish (pt) & 2,894 & 1,000 & 1,000 & 4,894 \\
Russian (ru) & 2,804 & 1,000 & 1,000 & 4,804 \\
French (fr) & 2,776 & 1,000 & 1,000 & 4,776 \\
Italian (it) & 2,589 & 1,000 & 1,000 & 4,589 \\
Korean (ko) & 2,452 & 1,000 & 1,000 & 4,452 \\
English (en) & 2,426 & 1,000 & 1,000 & 4,426 \\
Japanese (ja) & 2,000 & 1,000 & 1,000 & 4,000 \\
Arabic (ar) & 2,000 & 1,000 & 1,000 & 4,000 \\
Polish (pl) & 2,000 & 1,000 & 1,000 & 4,000 \\
Chinese-Traditional (zh-t) & 2,000 & 1,000 & 1,000 & 4,000 \\
Farsi (fa) & 2,000 & 1,000 & 1,000 & 4,000 \\
Indonesian (id) & 2,000 & 1,000 & 1,000 & 4,000 \\
Turkish (tr) & 2,000 & 1,000 & 1,000 & 4,000 \\
Vietnamese (vi) & 2,000 & 1,000 & 1,000 & 4,000 \\
Hebrew (he) & 1,807 & 1,000 & 1,000 & 3,807 \\
Romanian (ro) & 1,441 & 1,000 & 1,000 & 3,441 \\
Swedish (sv) & 1,419 & 1,000 & 1,000 & 3,419 \\
Filipino (tl) & 1,294 & 1,000 & 1,000 & 3,294 \\
Malay (ms) & 0 & 0 & 1,000 & 2,668 \\
Norwegian (no) & 0 & 0 & 1,000 & 1,098 \\
Catalan (ca) & 0 & 0 & 1,000 & 1,002 \\
Croatian (hr) & 0 & 0 & 907 & 907 \\
Georgian (ka) & 0 & 0 & 863 & 863 \\
Chinese-Simplified (zh-s) & 0 & 0 & 833 & 833 \\
Hungarian (hu) & 0 & 0 & 811 & 811 \\
Albanian (sq) & 0 & 0 & 809 & 809 \\
Serbian-Latin (sr-l) & 0 & 0 & 774 & 774 \\
\hline
\textbf{Total} & \textbf{47,437} & \textbf{21,000} & \textbf{28,997} & \textbf{99,202} \\
\hline
\end{tabular}
}
\caption{\label{GEMV-table} Language distribution and data statistics of GEM-V for multilingual video-language tasks. }
\end{table}

To the best of our knowledge, GEM dataset is the first multilingual vision-language dataset constructed for image-language and video-language tasks as the same time. GEM-I contains 1.2 million \{\textit{Query, Image, Title}\} triplets in 20 different languages for text-to-image retrieval and image captioning tasks. GEM-V contains 99K \{\textit{Query, Video, Title}\} triplets in 30 languages for text-to-video retrieval and video captioning tasks. In both GEM-I and GEM-V, \textit{Title} denotes the title of the web page where each image (or video) is extracted. This signal can be used as the auxiliary information in all GEM tasks, as it is usually highly relevant to the corresponding image (or video). 

%We provide baseline models for scenarios both with and without context present in Section \ref{sec:baseline-models}.

Next, we will describe how GEM-I and GEM-V are collected from a commercial search engine.

\subsection{GEM-I Construction}

First, we collect several billion images with Creative Commons licenses from the Internet, and discard images that contain pornographic or racy content. We also discard images with human faces, to avoid revealing privacy or introducing bias to our data. Besides, we only keep images which are larger than 300$\times$300 pixels to guarantee high image quality. The pornographic classifier, racy classifier, and human face classifier are trained and evaluated on human-labeled data. The (precision, recall) of them are (0.85, 0.92), (0.79, 0.94), and (0.85, 0.92), respectively.

Then, we collect user queries from a commercial search engine for each image based on user historical clicks. We also collect the title of the Web page that contains the image as the additional context, forming \{\textit{Query, Image, Title}\} triplets. Some text cleanup work is done to only keep high quality queries and contexts, including removing pornographic words and meaningless strings, and discarding very short queries or titles in that they are less likely to depict the image content, etc. We also apply an in-house GBDT model to filter out potentially highly irrelevant \{\textit{Query, Image, Title}\} triplets, which is trained using a small amount of human-labeled data, to predict the similarity between each \{\textit{Query}\} and \{\textit{Image, Title}\} pair.

Finally, we only keep the top 20 languages which have more than 1000 images, and sample 1.2 million \{\textit{Query, Image, Title}\}  triplets in total. The average length of query in GEM-I is 5.5 terms, which is shorter than 10.6 in MSCOCO \cite{chen2015microsoft} and 12.3 in Flicker30K \cite{vinyals2015tell}. Also, the average length of title is 10.1 terms. This makes GEM-I a more practical benchmark, since all data in GEM-I come from the real world, where the language configuration truly differs from the queries in existing datasets. For example, the queries in GEM were shorter and more concise, without perfect grammar or syntax structure. This makes GEM queries more "natural". Therefore, our benchmark can evaluate the models on data closer to real-world scenarios, so that the performance of the models will be more convincing in terms of being used in real-world applications. Based on human assessment on sampled query-image pairs, 83\% of the them are well matched pairs in that the query is a plausible caption of its paired image. We randomly split the data into train, dev and test sets within each language. The data statistics and language distribution of GEM-I can be found in Table~\ref{GEMI-table}. Figure~\ref{GEMI-case} gives some examples.

\subsection{GEM-V Construction}

We collect several billion videos from the Internet, and discard videos with pornographic or racy contents. We also discard very long videos to save storage and transfer expenses. For each video, its query and title are collected from a commercial search engine and cleaned-up according to a similar process as we described in GEM-I, where another in-house model is trained to filter out potentially irrelevant \{\textit{Query, Video, Title}\} triplets. 

\begin{figure*} [tp]
    \centering
    \includegraphics[width=\textwidth,height=\textheight,keepaspectratio]{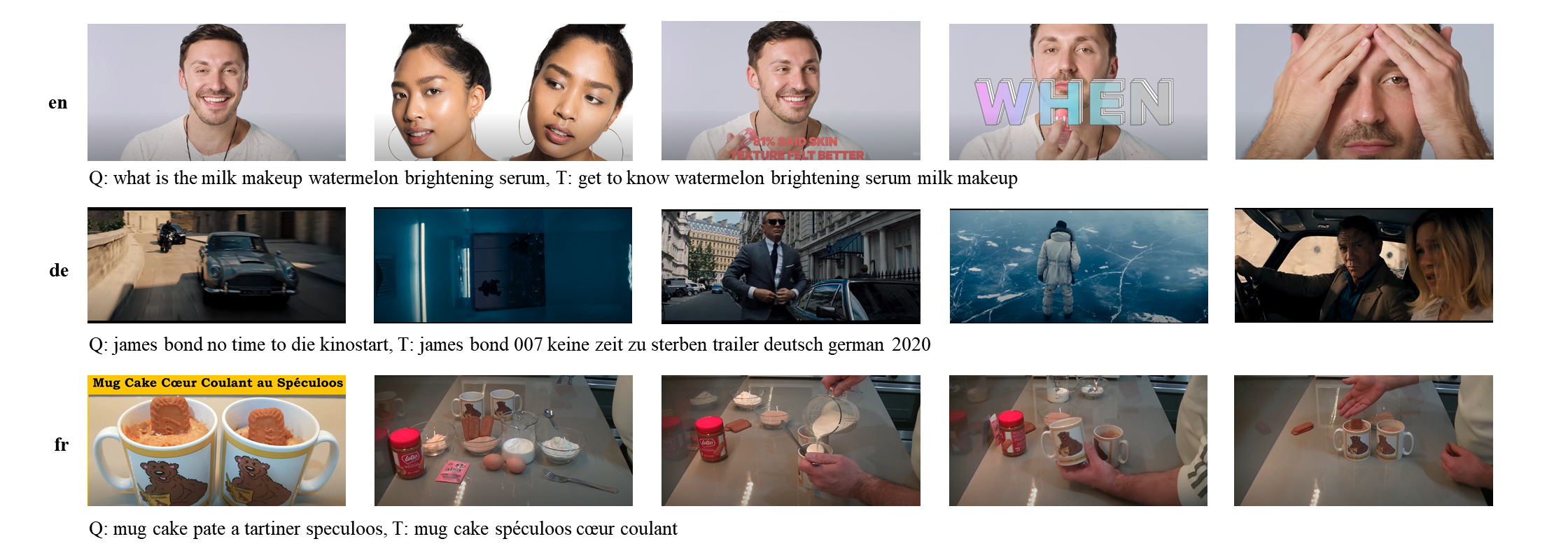}
    \caption{Examples in GEM-V dataset: Q: Query, T: Title.}
    \label{GEMV-case}
\end{figure*}

Finally, we only keep the top 30 languages which have more than 700 videos, and sample 99K \{\textit{Query, Video, Title}\} triplets in total. The total video length of GEM-V is 2,049 hours, and the average video length is 1.2 minutes. The average length of query in GEM-V is 5.3 terms, and that of title is 8.5 terms. We also conduct human evaluation on some sampled query-video pairs, and find 70\% of them are plausible matched pairs. We randomly split the data into train, dev and test sets within each language. The data statistics and language distribution of GEM-V can be found in Table~\ref{GEMV-table}. Figure~\ref{GEMV-case} gives some examples.

\section{Baseline Models}
\label{sec:baseline-models}

This section will introduce two baseline models for GEM, including M$^3$P, which is a multilingual multimodal pre-trained model for image-language tasks, and m-UniVL, which is a multilingual extension of UniVL \cite{luo2020univl} for video-language tasks.

\subsection{M$^3$P as Baseline of GEM-I}
We select M$^3$P \cite{ni2021m3p} as the baseline model for tasks in GEM-I, as it is the state-of-the-art multilingual image-language pre-trained model for both image-language understanding and generation tasks.

The M$^3$P model uses the model architecture of BERT for understanding tasks and a BERT-based encoder-decoder architecture for generation tasks. 
For understanding tasks, multilingual masked language modeling, multimodal masked language modeling, masked region modeling and visual-linguistic matching are used as pre-training tasks to train a Transformer-based encoder.
For generation tasks, multilingual denoising auto-encoding, image captioning and denoising image captioning are used as pre-training tasks to train a Transformer-based encoder-decoder. By training the encoder and the encoder-decoder with a multitask learning framework, universal representations are learned to map objects occurred in different modalities or expressed in different languages to vectors in a common semantic space.

\textbf{Fine-tune Tasks} Based on the pre-trained M$^3$P model, we further finetune it on our GEM-I data. For the text-image retrieval task, we adopt the BCE loss and NCE loss \cite{pmlr-v9-gutmann10a} (with equal weights) to learn the instance-level alignment between texts and images. The negative samples are generated by randomly forming text-image pairs from different training samples in the same batch. For the image captioning task, we directly learn captioning loss on GEM-I data.

\textbf{Side-Information}
Since title is considered as the side-information of the image, we concatenate it together with the image and feed them into the model. During the negative sampling process in the retrieval task, we treat title and image as a whole, i.e., for a certain query, the titles and images from other samples are considered as negatives.

\subsection{m-UniVL as Baseline of GEM-V}
We adopt the same model structure with the unified video and language pre-trained model UniVL \cite{luo2020univl}, which can perform both multi-modal understanding and generation tasks. Specifically, we extend the pre-trained UniVL model from monolingual to multilingual by replacing the BERT-based module with XLM-R \cite{DBLP:journals/corr/abs-1911-02116}, and call the new model m-UniVL. m-UniVL adopts an encoder-decoder architecture, including two single-modal encoders to encode the multilingual text and the visual features respectively, and one cross-modal encoder to learn the interactions between the two modalities, and finally an optional decoder for generation tasks. To better leverage the pre-trained models, we initialize each module with different pre-trained weights: for the multilingual text encoder, we directly initialize it with the pre-trained XLM-R\footnote{https://huggingface.co/xlm-roberta-base} \cite{DBLP:journals/corr/abs-1911-02116}, and for other modules including the visual encoder, the cross encoder and the decoder, we initialize them with the weights of the pre-trained UniVL\footnote{https://github.com/microsoft/UniVL}. 

\textbf{Fine-tune Tasks} Based on the pre-trained m-UniVL, we further finetune it using GEM-V data. For the text-video retrieval task, we only employ encoders in m-UniVL in the finetuning stage and use them to predict the matching score between text and video. There are two baseline models for the retrieval task: 1) m-UniVL(loose), the loosely coupled model that only uses the single-modal encoders; 2) m-UniVL(tight), the loosely plus tightly coupled model that includes both the single-modal encoders and the cross-modal encoder. We adopt the NCE loss \cite{pmlr-v9-gutmann10a} to learn to discriminate the positive video-text pairs against negative ones. The negative video-text samples are created by replacing the text or video in a positive sample with randomly-selected text or video from other samples. For video captioning task, we employ all modules including all the encoders and decoder to learn the caption generation task.

\textbf{Side-Information} In regard to the titles, we use them as the side-information of the videos for an efficient text-video retrieval. In details, we first map the embedding of the title to the same dimension with the video embedding, and then concatenate them together. Then we encode them using the visual encoder to generate the enhanced video features for further processing.
%This two-stream loosely coupled model benefits retrieval task considering calculation efficiency. 

\section{Related Work}

\subsection{Natural Language Benchmarks}
GLUE \cite{DBLP:journals/corr/abs-1804-07461} and SuperGLUE \cite{DBLP:journals/corr/abs-1905-00537} are two comprehensive datasets that can be used to train and evaluate natural language understanding systems. GLGE \cite{liu2020glge} is another comprehensive dataset for natural language generation evaluation. XGLUE \cite{liang2020xglue} and XTREME \cite{hu2020xtreme} are two recent benchmark efforts that extend the evaluation scenarios from monolingual to multilingual. Recent pre-trained language models benefit a lot from these datasets, by evaluating their effectiveness under a relatively fair environment.

\subsection{Vision-Language Benchmarks} 

A number of vision-language datasets have been widely used in the multimodal research. 

MSCOCO \cite{chen2015microsoft} and Flicker30K \cite{vinyals2015tell} are two datasets for image-text retrieval and image-captioning tasks. These two benchmarks have been extended to multilingual tasks \citep{elliott2016multi30k,elliott2017findings,mcoco1,mcoco2} as well. VQA \cite{VQA} and GQA \cite{hudson2018gqa} are two datasets for visual question answering. VCR \cite{DBLP:journals/corr/abs-1811-10830} is another dataset for visual commonsense reasoning. Comparing with all these existing datasets, GEM-I has unique characteristics. First, it is a large-scale multilingual image-text dataset covering 20 different languages. Second, the query-image pairs in GEM-I come from a commercial search engine. Therefore, it has big practical values. Third, for each query-image pair, the title of the Web page that contains the image is also included as the additional context, which makes GEM-I different from all existing datasets.

HowTo100M \cite{DBLP:journals/corr/abs-1906-03327}, YouCook2 \cite{ZhXuCoCVPR18}, and MSR-VTT \cite{xu2016msr-vtt} are three typical benchmarks for video-text retrieval and video captioning tasks. TVQA \cite{DBLP:journals/corr/abs-1809-01696} and ActivityNet-QA \cite{DBLP:journals/corr/abs-1906-02467} are two typical benchmarks for video question answering. Comparing with all these existing datasets, GEM-V is the first video-language benchmark supporting multilingual scenarios. Similar to GEM-I, it also has big practical values, as all data in GEM-V come from a real-world search engine with massive users.

\section{Experiments}
In this section, we evaluate two baseline pre-trained models (described in Section~\ref{sec:baseline-models}) on GEM. Specifically, M$^3$P is evaluated on GEM-I for multilingual image-language tasks and m-UniVL is evaluated on GEM-V for multilingual video-language tasks. For both baseline models, we fine-tune them on downstream tasks directly.

%%%%%%%%%%%%%%%%%%%%%%%%%%%%%%%%%%%%%%%%%%%%%%%%%%%%%%%%%%%%%%%%%%%%%%%
\subsection{Image-Language Evaluation on GEM-I}

\begin{table*}[tp] 
	\centering
	\scalebox{0.64}{
		\begin{tabular}{l|cccccccccc|c}
			\toprule 
			Setting                     &en     &es    &fr    &it    & pt   &de    &ko    &pl    &ca    &nl   &-\\ 
			\midrule 
			% Q-I pairs  & &  &  &  &  &  &  &   \\
			% \midrule 
			\multicolumn{11}{c}{\textit{Zero-Shot}} \\
			M$^3$P (Q$\rightarrow$I)       &22.08  &8.31  &8.51  &7.16  &7.05  &9.31  &4.68  &4.38  &9.10  &9.09 &-\\ 
			M$^3$P (Q$\rightarrow$I+T)     &5.80  &4.16  &3.98  &2.9   &3.71  &3.36  &2.68  &4.07  &2.86  &3.96 &-\\
			\multicolumn{11}{c}{\textit{Fine-tune on ALL}} \\
			M$^3$P (Q$\rightarrow$I)     	&43.85 &26.15 &24.83 &22.72 &27.05 &27.18 &15.80 &32.80 &12.83 &20.78 &-\\
			M$^3$P (Q$\rightarrow$I+T)     &93.75 &93.16 &95.15 &93.26 &93.73 &89.37 &67.46 &82.67 &90.78 &90.37 &-\\
			% M3P (image captioning fine-tune on all)   &   &- & - & - & - & - & - & - & - \\
			\midrule \midrule
			                            &ja     &id    &vi    &cs    &ro    &tr &gl & hr & hu & ms &AVG\\ 
			\midrule
			\multicolumn{11}{c}{\textit{Zero-Shot}} \\
			M$^3$P (Q$\rightarrow$I) 		&7.68 &12.06 &5.00   &5.32  &5.81  &5.13 &4.72 &5.30 &4.30 &8.08 &7.65\\ 
			M$^3$P (Q$\rightarrow$I+T)     &4.5   &4.53  &2.08  &3.72  &3.61  &3.28  &2.55  &2.58  &3.26  &3.41 &3.55\\ 
			\multicolumn{11}{c}{\textit{Fine-tune on ALL}}\\
			M$^3$P (Q$\rightarrow$I)   	&18.88 &22.13 &10.10 &13.33 &16.10 &13.23 &10.38 &13.51 &11.11 &14.33 &19.85 \\
			M$^3$P (Q$\rightarrow$I+T)    	&71.90 &81.98 &54.83 &76.23 &64.35 &71.97 &75.00 &71.28 &63.43 &74.18 &79.74\\
			
% 			\\https://www.overleaf.com/project/600551c9137957edbb4777cc
			% M3P (image captioning fine-tune on all) &     &- & - & - & - & - & - & - & - \\
			% \midrule
			% \midrule
			% Methods                                  &input    &en     &es    &fr    &it    & pt   &de    &ko    &pl    &ca    &nl   \\ https://www.overleaf.com/project/600551c9137957edbb4777cc
			% \midrule 
			% M3P (zero-shot image retrieval)          &\multirow{3}{*}{Q-I-C pair}     &5.80  &4.16 &3.98  &2.9  &3.71  &3.36  &2.68  &4.07  &2.86  &3.96 \\ 
			% M3P (image retrieval fine-tune on all)   &    &93.75 & 93.16 & 95.15 & 93.26 & 93.73 & 89.37 & 67.46 & 82.67 &90.78&90.37\\
			% M3P (image captioning fine-tune on all)  &    &- & - & - & - & - & - & - & - \\
			% \midrule
			%                                 		&input	 &ja     &id    &vi    &cs    &ro    &tr &gl & hr & hu & ms\\ 
			% \midrule
			% M3P (zero-shot image retrieval) 		 &\multirow{3}{*}{Q-I-C pair} &4.5 &4.53 &2.08   &3.72  &3.61 &3.28 &2.55 &2.58 &3.26 &3.41\\ 
			% M3P (image retrieval fine-tune on all)   &    &71.90 & 81.98 & 54.83 & 76.23 & 64.35 & 71.97 & 75.00 & 71.28&63.43&74.18 \\
			% % M3P (image captioning fine-tune on all)  &    &- & - & - & - & - & - & - & - \\

			\bottomrule
		\end{tabular}
	}
	\caption{Evaluation results of M$^3$P on GEM-I test set for text-to-image retrieval tasks where Mean-Recall is taken as metric. \textbf{Q$\rightarrow$I} denotes
	the setting where only image (I) is used to compute its similarity with query (Q), \textbf{Q$\rightarrow$I+T} denotes
	the setting where both image (I) and title (T) are used to compute the similarity with query (Q). The average score is computed over all 20 languages.} 
	\label{tab:result_of_image_text_retrieval}
\end{table*}

\begin{table*}[tp] 
	\centering
	\scalebox{0.64}{
		\begin{tabular}{l|l|cccccccccc|c}
			\toprule 
			Setting & Metric           &en    &es    &fr    &it    &pt   &de    &ko    &pl    &ca    &nl  &- \\ 
			\midrule 
			M$^3$P  & ROUGE-L   	   &6.97  &13.86 &10.47 &9.13  &8.35 &8.67 & 3.27 & 9.31 &12.84  &3.62 &- \\
			M$^3$P  & METEOR   	   &3.21  &5.84 &4.91 &4.14  &3.67 &3.81 & 2.21 & 4.01 &5.7  &1.54 &- \\
			M$^3$P  & CIDEr   	   &17.89  &8.68 &14.92 &11.98  &7.93 &20.18 & 7.33 & 8.44 &7.59  &6.79 &- \\
			\midrule \midrule 
			                       &ja    &id    &vi    &cs    &ro   &tr &gl & hr & hu & ms &AVG\\ 
			\midrule
			M$^3$P 	& ROUGE-L          &4.10  &0.96 & 0.66 & 4.58 & 3.98 &0.32  &9.84&0.25 &0.57&0.39 &5.61\\
			M$^3$P 	& METEOR           &1.26  &0.47 & 0.22 & 1.97 & 1.78 &0.15  &4.43&0.11 &0.26&0.18 &2.49\\
			M$^3$P 	& CIDEr            &5.01  &2.03 & 1.14 & 5.72 & 2.96 &0.79  &6.86&0.72 &1.43&0.91 &6.97\\
			\bottomrule
		\end{tabular}
	}
	\caption{Evaluation results of M$^3$P on GEM-I test set for image captioning task where ROUGE-L, METEOR and CIDEr are taken as metrics. Only images (without title) are used to test its multilingual multimodal captioning ability. The average score is computed over all 20 languages.}
	\label{tab:result_of_image_captioning}
\end{table*}

\subsubsection{Experimental Settings}

We select the open-source version\footnote{https://github.com/microsoft/M3P} of M$^3$P \cite{ni2021m3p} for the image-language evaluation on GEM-I. It uses 101G sentences (in 100 languages) extracted from Wikipedia as the multilingual pre-training corpus, and uses 3.3 million English image-caption pairs in Conceptual Captions \cite{sharma2018conceptual} as the multimodal pre-training corpus. 

For text-to-image retrieval, the hyper-parameters of the encoder are set as follows: 768 hidden units, 12 heads, GELU activation, a dropout rate of 0.1, 128 max input length, 12 layers in encoder. For image captioning, the hyper-parameters of the encoder-decoder are set as follows: 768 hidden units, 8 heads, GELU activation, a dropout rate of 0.1, 128 max input length, 12 layers in both encoder and decoder. The transformer parameters between the encoder and decoder are shared, including embedding modules and self-attention modules.

We fine-tune M$^3$P on text-to-image retrieval and image captioning tasks. For retrieval, we use Adam optimizer with $\beta_1=0.9$, $\beta_2=0.98$, an initial learning rate of 5e-5, a weight decay of 1e-4 and a batch size of 64 to fine-tune M$^3$P for 30 epochs. For captioning, a learning rate of 1e-4 and a batch size of 16 are used to fine-tune M$^3$P for 20 epochs. All above calculations are carried on 4 NVIDIA Tesla P100 GPUs. 
% To test the GEM-I dataset, we take M3P~\cite{ni2021m3p} model 
% which is pretrained on large multilingual corpus and multimodal 

%For image encoding, we take an external object detector~\cite{ren2015faster} pretrained on Visual Genome dataset~\cite{krishna2017visual} to prepare at most 100 region features for each image. Simultaneously, a pre-trained XLM-R model with subword tokenization is employed for text embedding.

%We utilize a standard Transformer to encode multimodal feature representation, which consists of 12 layers of blocks with each blocking having a hidden state size of 768 and 12 attention heads.

\subsubsection{Text-to-Image Retrieval Results}
We follow the same evaluation metric,  mean-Recall (average score of R@1, R@5, R@10), in M$^3$P to report its the performance on text-to-image retrieval task on GEM-I dataset. 

From the results reported in Table \ref{tab:result_of_image_text_retrieval}, 
we have several observations:

1) When applying M$^3$P to GEM-I without fine-tuning (i.e. the zero-shot setting), the general performance is poor. The major reason is that M$^3$P is pre-trained on a monolingual multimodal corpus and a multilingual monomodal corpus, and both datasets have very different data distributions comparing with GEM-I.

2) By fine-tuning M$^3$P using all labeled data in different languages (i.e. the fine-tune on all setting), better performance can be obtained. This shows the strong transfer ability of M$^3$P, when there is a moderate amount of labeled data for fine-tuning.

3) By furthering considering the title signal in this retrieval task, the general performance can be improved significantly. This indicates the strong correlation between the query and the title. Besides, when taking the title signal into the zero-shot setting, we can observe a performance drop. It is due to that M$^3$P is pretrained with input paradigm Q-I, thus making it not suitable for evaluating Q-I-T paradigm directly.
%% the reason(I guess) for why injecting title into zero-shot setting performs worse

%the M$^3$P model achieves xxx\% mean-Recall in English language but performs relatively poor in other languages. This is mainly caused by long-tailed distribution in GEM-I datasets and most of image-text pairs are English, which results in imbalance fine-tuning when learning multilingual cross-modal representation. It leaves a big room to investigate imbalance fine-tuning in GEM-I dataset for future research. Furthermore, we also demonstrate the results of zero-shot text-based image retrieval to study the multilingual multimodal knowledge captured in M$^3$P model. We evaluate GEM-I test set by using M$^3$P pre-trained model directly as in Table~\ref{tab:result_of_image_text_retrieval}. We can see that M$^3$P has a certain capacity to capture unified representation among different languages.

\subsubsection{Image Captioning Results}
As in Table~\ref{tab:result_of_image_captioning}, we report the performance of image captioning tasks on GEM-I test set with M$^3$P model where ROUGE-L \cite{lin-och-2004-automatic}, METEOR \cite{banerjee2005meteor} and CIDEr \cite{vedantam2015cider} are taken as the evaluation metrics. To study the image captioning ability of M$^3$P, we only use images (without title) to generate queries in GEM-I dataset. In general, the M$^3$P model performs relatively poor on this task, due to that most search queries are short keywords instead of a complete sentence, and they differ from our pre-training data a lot. 

From the above results from text-to-image retrieval task and the image captioning task, we can conclude that our proposed GEM-I dataset can demonstrate a model's image understanding and generation ability in 20 different languages.

%%%%%%%%%%%%%%%%%%%%%%%%%%%%%%%%%%%%%%%%%%%%%%%%%%%%%%%%%%%%%%%%%%%%%%%%%

\subsection{Video-Language Evaluation on GEM-V}
\begin{table*}[tp] 
	\centering
	\scalebox{0.63}{
		\begin{tabular}{l|ccccccccccccccc|c}
			\toprule 
			Setting & en & id & pt & vi & ro & ko & ja & fr & ar & de & tl & sv & fa & he & it &- \\ 
			\midrule 
			\multicolumn{17}{c}{\textit{Fine-tune on ALL}} \\
			m-UniVL(loose) (Q$\rightarrow$V) & 23.27 & 17.67 & 23.50 & 12.83 & 19.90 & 12.83 & 12.37 & 24.77 & 9.03 & 22.27 & 11.23 & 16.07 & 9.87 & 9.50 & 23.60 &-\\  
			%m-UniVL(loose) (query\#tag$\rightarrow$video) & 23.73 & 17.33 & 22.37 & 12.50 & 19.20 & 13.63 & 12.87 & 24.63 & 8.80 & 22.27 & 10.10 & 15.58 & 10.27 & 9.97 & 21.27 \\  
			%m-UniVL(loose) (query\#tag$\rightarrow$title) & 83.83 & 73.30 & 74.32 & 57.50 & 68.20 & 59.53 & 60.23 & 77.25 & 54.60 & 69.53 & 60.91 & 54.17 & 58.24 & 61.25 & 70.03 \\
			m-UniVL(loose) (Q$\rightarrow$V+T) & 71.83 & 57.03 & 62.90 & 38.97 & 53.57 & 42.09 & 38.87 & 62.83 & 36.50 & 56.67 & 43.37 & 40.20 & 39.67 & 35.20 & 56.80  &-\\
			m-UniVL(tight) (Q$\rightarrow$V+T) & 83.87 & 69.97 & 61.20 & 52.07 & 59.08 & 63.17 & 66.10 & 74.47 & 46.92 & 70.27 & 59.23 & 51.60 & 58.10 & 56.60 & 58.90  &-\\
			\multicolumn{17}{c}{\textit{Fine-tune on EACH}} \\
			m-UniVL(loose) (Q$\rightarrow$V+T) & 28.90 & 19.97 & 21.23 & 12.33 & 14.93 & 16.87 & 13.30 & 20.73 & 10.57 & 21.70 & 14.60 & 12.23 & 12.03 & 10.27 & 18.70  &-\\
			\midrule \midrule
			& tr & ru & nl & pl & zh-t & es & ms & no & ca & hr & ka & zh-s & hu & sq & sr-l  &AVG\\ 
			\midrule
			\multicolumn{17}{c}{\textit{Fine-tune on ALL}} \\
			m-UniVL(loose) (Q$\rightarrow$V) & 18.70 & 18.10 & 20.00 & 22.50 & 16.00 & 25.13 & 8.10 & 16.70 & 9.05 & 13.38 & 7.67 & 9.64 & 11.56 & 6.80 & 11.03 &15.44\\  
			%m-UniVL(loose) (query\#tag$\rightarrow$video) & 18.47 & 18.90 & 19.80 & 25.50 & 16.57 & 24.67 & 9.13 & 18.70 & 9.11 & 13.49 & 7.90 & 10.20 & 9.80 & 8.16 & 11.54 \\ 
			%m-UniVL(loose) (query\#tag$\rightarrow$title) & 66.17 & 72.60 & 66.34 & 63.57 & 67.00 & 79.33 & 67.46 & 55.99 & 67.25 & 53.88 & 49.40 & 57.70 & 63.47 & 52.00 & 55.78 \\
			m-UniVL(loose) (Q$\rightarrow$V+T) & 51.43 & 57.73 & 49.73 & 49.80 & 45.70 & 66.03 & 43.97 & 40.60 & 41.40 & 38.37 & 28.00 & 38.38 & 40.69 & 33.00 & 35.70 &46.57\\
			m-UniVL(tight) (Q$\rightarrow$V+T) & 63.13 & 70.57 & 59.07 & 50.67 & 69.10 & 64.97 & 64.60 & 54.40 & 59.37 & 31.50 & 39.05 & 64.31 & 58.73 & 50.97 & 32.86 &58.83 \\
			\multicolumn{17}{c}{\textit{Fine-tune on EACH}} \\
			m-UniVL(loose) (Q$\rightarrow$V+T) & 21.00 & 17.70 & 21.70 & 19.50 & 17.40 & 19.83 & \textit{NA} & \textit{NA} & \textit{NA} & \textit{NA} & \textit{NA} & \textit{NA} & \textit{NA} & \textit{NA} & \textit{NA} &17.40\\
			\bottomrule
		\end{tabular}
	}
	\caption{Evaluation results of m-UniVL on GEM-V for text-to-video retrieval, where Mean-Recall is used as the metric. \textbf{Q$\rightarrow$V} denotes the setting where only video (V) is used to compute its similarity with query (Q). \textbf{Q$\rightarrow$V+T} denotes the setting where both video (V) and title (T) are used to compute their similarity with query (Q). The average score is computed over all 30 languages.  
	}
	\label{tab:result_of_video_retrieval}
\end{table*}
\subsubsection{Experimental Settings}
We select the open-source version\footnote{https://github.com/microsoft/UniVL} of UniVL \cite{luo2020univl} and replace the original text encoder with XLM-R \cite{conneau-etal-2020-unsupervised}, to support the multilingual video-language evaluation on GEM-V. The original UniVL is pre-trained on 1.2 million instructional videos with ASR transcripts in HowTo100M \cite{DBLP:journals/corr/abs-1906-03327}.

\begin{table*}[tbp] 
	\centering
	\scalebox{0.54}{
		\begin{tabular}{l|l|ccccccccccccccc|c}
			\toprule
			Setting & Metric & en & id & pt & vi & ro & ko & ja & fr & ar & de & tl & sv & fa & he & it &- \\ 
			\midrule
			\multicolumn{17}{c}{\textit{Fine-tune on ALL}} \\
			m-UniVL (V$\rightarrow$Q) & ROUGE-L   & 9.43 & 10.95 & 14.63 & 6.64 & 9.96 & 3.41 & 3.10 & 16.87 & 3.80 & 8.18 & 9.72 & 8.55 & 8.25 & 2.28 & 10.65 &-\\ 
			m-UniVL (T$\rightarrow$Q) & ROUGE-L        & 46.06 & 35.89 & 36.68 & 27.36 & 35.08 & 21.01 & 17.14 & 41.64 & 22.40 & 30.30 &43.36 & 26.10 & 30.58 & 33.07 & 34.43 &-\\ 
			m-UniVL (V+T$\rightarrow$Q) & ROUGE-L      & 46.01 & 36.73 & 37.59 & 26.75 & 36.41 & 20.71 & 17.12 & 41.90 & 23.45 & 30.24 & 44.40 & 26.38 & 30.21 & 32.82 & 34.98  &- \\
			
			m-UniVL (V$\rightarrow$Q) & METEOR   & 3.93 & 4.51 & 6.09 & 2.39 & 4.09 & 2.99 & 3.38 & 8.06 & 15.16 & 3.71 & 4.31 & 3.62 & 17.06 & 12.04 & 4.65 &-\\ 
			m-UniVL (T$\rightarrow$Q) & METEOR         & 23.99 & 17.01 & 17.40 & 13.13 & 17.10 & 20.77 & 18.33 & 21.25 & 26.47 & 14.40 & 22.67 & 12.23 & 26.25 & 29.33 & 16.60 &-\\ 
			m-UniVL (V+T$\rightarrow$Q) & METEOR       & 23.98 & 17.44 & 18.01 & 12.52 & 18.09 & 20.34 & 18.24 & 21.39 & 26.76 & 14.52 & 23.00 & 12.26 & 26.17 & 28.91 & 16.94 &- \\
			
			m-UniVL (V$\rightarrow$Q) & CIDEr  & 19.16 & 18.35 & 23.58 & 13.53 & 18.56 & 7.12 & 4.20 & 43.82 & 5.72 & 18.23 & 23.40 & 8.99 & 14.88 & 4.75 & 17.53 &-\\ 
			m-UniVL (T$\rightarrow$Q) & CIDEr          & 256.65 & 155.05 & 164.26 & 116.46 & 164.89 & 74.95 & 57.15 & 223.09 & 86.57 & 138.50 & 220.58 & 101.64 & 119.49 & 139.39 &170.02 &-\\
			m-UniVL (V+T$\rightarrow$Q) & CIDEr        & 255.04 & 157.64 & 167.79 & 108.66 & 174.59 & 74.01 & 58.73 & 223.67 & 90.70 & 138.66& 223.21 & 100.61 & 115.37 & 136.24  &174.44 &- \\

			\multicolumn{17}{c}{\textit{Fine-tune on EACH}} \\
			m-UniVL (V+T$\rightarrow$Q) & ROUGE-L      & 21.30 & 12.25 & 18.36 & 6.45 & 9.09 & 7.59 & 3.77 & 15.20 & 4.92 & 8.94 & 14.92 & 8.11 & 7.31 & 1.76 & 10.05 &-\\
			m-UniVL (V+T$\rightarrow$Q) & METEOR       & 9.50 & 4.92   & 7.99  & 2.44 & 3.67 & 6.79 & 3.48 & 7.01 & 17.21 & 3.92 & 7.12 & 3.47 & 16.66 & 12.75 & 4.22 &-\\
			m-UniVL (V+T$\rightarrow$Q) & CIDEr        &65.52 & 24.26 & 44.69 & 15.47 & 21.45 & 23.40 & 9.50 & 43.71 & 8.35 & 21.32 & 41.04 & 9.89 & 13.65 & 3.71 & 18.12 &-\\
			
			\midrule \midrule
			& tr & ru & nl & pl & zh-t & es & ms & no & ca & hr & ka & zh-s & hu & sq & sr-l &AVG\\ 
			\midrule
			m-UniVL (V$\rightarrow$Q) & ROUGE-L      & 11.53 & 14.44& 9.83 & 14.19 & 3.95 & 18.07 & 0.15 & 0.91 & 0.27 & 2.49 & 0.00 & 0.11 & 2.68 & 1.29 & 0.33 &6.89\\
			m-UniVL (T$\rightarrow$Q) & ROUGE-L      & 32.19 & 35.41 & 27.77 & 29.99 & 15.72 & 41.35 & 34.44 & 21.35 & 38.29 & 24.94 & 1.70 & 4.63 & 30.13 & 27.22 & 28.28 &29.15\\ 
			m-UniVL (V+T$\rightarrow$Q) & ROUGE-L      & 34.43 & 35.84 & 28.31 & 29.82 & 15.50 & 42.29 & 36.36 & 21.59 & 38.09 & 26.09 & 2.92 & 4.55 & 30.58 & 28.37 & 29.51 &29.67  \\
			
			m-UniVL (V$\rightarrow$Q) & METEOR 		& 5.41 & 6.11 & 4.47 & 6.06 & 2.65 & 8.21 & 0.07 & 0.39 & 0.13 & 1.05 & 0.00 & 0.61 & 1.18 & 0.67 & 0.13 & 4.44 \\
			m-UniVL (T$\rightarrow$Q) & METEOR  	 &15.53 & 16.79 & 13.21 & 13.91 & 16.03 & 20.50 & 16.88 & 9.77 & 19.21 & 12.00 & 1.74& 12.24 & 13.66 & 13.17 & 13.53 & 16.84\\		
			m-UniVL (V+T$\rightarrow$Q) & METEOR      &16.54 & 17.12 & 13.62 & 13.86 & 15.48 & 20.83 & 17.92 & 9.83 & 18.95 & 12.47 & 2.76 & 11.59 & 13.78 & 13.72 & 13.81 & 17.03\\
			
			m-UniVL (V$\rightarrow$Q) & CIDEr 		& 28.84 & 24.05 & 17.35 & 17.37 & 8.00 & 32.89 & 0.15 & 1.65 & 0.45 & 2.20 & 0.00 & 0.29 & 2.71 & 1.70 & 0.73 & 12.67\\
			m-UniVL (T$\rightarrow$Q) & CIDEr  	     & 149.09 & 161.47 & 123.79 & 117.37 & 47.94 & 203.88 & 156.38 & 84.81 & 192.65 & 102.38 & 3.95 & 13.58 & 129.47 & 109.78 & 119.64 &130.16\\
			m-UniVL (V+T$\rightarrow$Q) & CIDEr      & 156.35 & 166.33 & 124.20 & 116.29 & 47.13 & 204.85 & 166.13 & 82.70 & 189.35 &106.02 & 8.01 & 12.85 & 128.94 & 114.98 & 118.01 & 131.38\\

			\multicolumn{17}{c}{\textit{Fine-tune on EACH}} \\
			m-UniVL (V+T$\rightarrow$Q) & ROUGE-L      & 11.46 & 17.52 & 9.16 & 13.31 & 3.15 & 24.21 & \textit{NA} & \textit{NA} & \textit{NA} & \textit{NA} & \textit{NA} & \textit{NA} & \textit{NA} & \textit{NA} & \textit{NA} &10.90\\
			m-UniVL (V+T$\rightarrow$Q) & METEOR      &5.20 & 7.44 & 3.93 & 5.65 & 2.08 & 11.19 & \textit{NA} & \textit{NA} & \textit{NA} & \textit{NA} & \textit{NA} & \textit{NA} & \textit{NA} & \textit{NA} & \textit{NA} &6.98\\
			m-UniVL (V+T$\rightarrow$Q) & CIDEr      &29.62 & 36.11 & 16.48 & 21.15 & 6.56 & 62.26& \textit{NA} & \textit{NA} & \textit{NA} & \textit{NA} & \textit{NA} & \textit{NA} & \textit{NA} & \textit{NA} & \textit{NA} &25.54\\

			\bottomrule
		\end{tabular}
	}
	\caption{Evaluation results of m-UniVL on GEM-V for video captioning, where ROUGE-L, METEOR and CIDEr are taken as metrics. \textbf{V$\rightarrow$Q} and \textbf{T$\rightarrow$Q} denote the video caption is generated based on video (V) and title (T), respectively. \textbf{V+T$\rightarrow$Q} denotes the video caption is generated based on both video (V) and title (T). The average score is computed over all 30 languages.
	}
	\label{tab:caption_result}
\end{table*}

For text-to-video retrieval task, m-UniVL extracts video features using the off-the-shelf pre-trained S3D \cite{DBLP:journals/corr/abs-1912-06430} model. The FPS of the 3D feature extractor is 16 and the dimension is 1,024. The hyper-parameters of the video encoder are set as follows: 768 hidden units, 12 heads, 6 layers of of Transformer blocks to capture the sequential information on the 3D features. The hyper-parameters of the text encoder are identical to the ones in XLM-R: 768 hidden units, 12 heads, 12 layers of Transformer blocks. The cross encoder on the top of the text and video encoders has 2 layers with 768 hidden units and 12 heads. For video captioning, the decoder is with 3 layers, 768 hidden units and 12 heads.

We finetune m-UniVL on text-to-video retrieval and video captioning tasks. For retrieval, a learning rate of 1e-4 and a batch size of 128 are used to fine-tune m-UniVL for 50 epochs. For captioning, a learning rate of 3e-5 and a batch size of 16 are used to fine-tune m-UniVL for 5 epochs. All above calculations are carried on 4 NVIDIA Tesla V100 GPUs.

%For text encoding, we employ the pre-trained XLM-R model, which adopts subword tokenization with Sentence Piece \cite{kudo-richardson-2018-sentencepiece}. We exploit the XLM-R$_\text{Base}$ model with 12 layers of Transformer blocks. Each block has 12 attention heads and the hidden size is 768.

\subsubsection{Text-to-Video Retrieval Results}
Following official UniVL on retrieval task, we evaluate the text-to-video retrieval task on our GEM-V using two variants. One is m-UniVL (loose), which encodes the input text query and candidate video clips (and optional title) through the text encoder and video encoder respectively and finally calculates the matching score through dot product. The other is m-UniVL (tight), based on m-UniVL (loose), m-UniVL (tight) further concatenates the encoded features and feeds them to the cross encoder to get unified representation and predict the matching score on the first token `$\langle$s$\rangle$'. The evaluation metric is mean-Recall (arithmetic mean of Recall@$K$ for $K\in \{1,5,10\}$). 

Tables \ref{tab:result_of_video_retrieval} lists the retrieval results. The results can be divided into two groups. One is from the fine-tuning on all training set across linguistic type (\textit{Fine-tune on ALL}), and the other is from the fine-tuning on individual training set of each language (\textit{Fine-tune on Each}). The target of such a division is to explore whether one language can benefit from other wide languages. Besides, there are 9 languages without training set. We keep such a zero-shot evaluation to explore the transfer ability of the proposed model.

We can get three conclusions from the results:

1) The m-UniVL (tight) outperforms m-UniVL (loose) at the same retrieval settings. It proves that the cross encoder of UniVL enables the multi-modality features to fully interact with each other to capture better alignment.

2) The title of the video introduces a large performance gain and is a good semantic feature of the video. This metadata is especially useful for zero shot setting with a significant improvement. They demonstrate that improving the retrieval performance on pure videos without titles is still a challenge. Our proposed GEM develops a chance to push such a multimodal challenge.

3) Fine-tune on all can achieve better results than fine-tune on each. The reason is the former can effectively leverage the data from all languages and benefit the task rather than the latter. Besides, for zero shot languages, fine-tune on all is also very effective. It demonstrates that our proposed GEM can also be used on multilingual research besides multimodal research.

\noindent	
\subsubsection{Video Captioning Results}
The captioning task aims to generate a caption given a video clip (and optional title) in our setting. Such a generation task is from our practical application. We adopt whole m-UniVL including encoders and decoder to finish the task. The evaluation metric are  ROUGE-L \cite{lin-och-2004-automatic}, METEOR \cite{banerjee2005meteor} and CIDEr \cite{vedantam2015cider}, whose values are obtained from an open-source tool\footnote{https://github.com/Maluuba/nlg-eval}.

Table \ref{tab:caption_result} lists the experimental results. Similar conclusions can be drawn as the retrieval task, and there are two more observations: 

1) For captioning task, the performance of the generation on pure videos is low. The reason is that the search queries sometimes are the keywords instead of a whole sentence, thus the task of V$\rightarrow$Q is relatively hard.

2) Title is especially important due to the characteristic of this data collection process.

From the above results from the text-to-video retrieval task and the video captioning task, we can conclude that our proposed GEM-V can improve video understanding and generation under the multilingual and multimodal perspective.

%\subsection{Ablation Study}

\section{Conclusion}
This paper presents GEM as a benchmark for evaluating the generalization capabilities of vision-language models on image-language and video-language tasks. GEM is also the first large-scale multilingual multimodal dataset, where the natural language contexts are collected from a commercial search engine in 20 and 30 languages for image-related and video-related tasks, respectively. We describe two vision-language pre-trained models for GEM and hope these efforts can advance the development of multilingual multimodal research.

\section{Ethical Considerations}

We have reviewed our data release process and it has been approved by our institutional review board. Specifically, (a) In GEM-I, all of the images are with proper Creative Commons Licenses, so that they are safe to be distributed without violating any policies or intellectual rights. Also, we discarded images with human faces to avoid revealing privacy. (b) In GEM-V, all of the videos were originated from Youtube, and we will only provide Youtube URLs to the researchers. We have confirmed with our institutional review board that distributing URLs does not violate any policies or intellectual rights. We didn't do anything specific for human faces in the videos, since we are only distributing video URLs, and modifying the original videos (such as blurring the faces) might violate the copyright of the videos. When releasing GEM to the public, we will indicate the data source, emphasize that the dataset is for research purposes only, and provide an email address for people to contact us to delete any data if any infringement. During data collection, we didn't collect, store, or distribute any private information of the users.

To measure the quality of our dataset, we employed crowd-sourcing judges in the United States and provided labeling guidelines for them. The compensation given to the workers is 15 USD per hour for GEM-I and 25 USD per hour for GEM-V. The level of compensation is determined by: (a) Market price according to similar labeling tasks in the US. (b) The difficulty and labeling speed of this task. This task involves labeling if a query is related to an image or video, so it is considered as a relatively easy task. The labeling speed is about 300 query-image pairs per hour and 180 query-video pairs per hour.

\bibliographystyle{acl_natbib}
\bibliography{anthology,acl2021}

%\appendix

\end{document}